# Intensity and Rescale Invariant Copy Move Forgery Detection Techniques

Tejas K, Swathi C, Rajesh Kumar M, *Senior Member, IEEE*
School of Electronics and Communication Engineering
Vellore Institute of Technology
Email: tejas.krishnareddy1415@gmail.com, chintala.swathi14@gmail.com
mrajeshkumar@vit.ac.in

*Abstract*—In this contemporary world, digital media such as videos and images behave as an active medium to carry valuable information across the globe on all fronts. However, there are several techniques evolved to tamper the image which has made their authenticity untrustworthy. Copy-Move Forgery (CMF) is one of the most common forgeries present in an image where a cluster of pixels are duplicated in the same image with potential post-processing techniques. Various state-of-art techniques are developed in the recent years which are effective in detecting passive image forgery. However, most methods do fail when the copied image is rescaled or added with certain intensity before being pasted due to de-synchronization of pixels in searching process. To tackle this problem, the paper proposes distinct novel algorithms which recognize a unique approach of using Hu's invariant moments and Discreet Cosine Transformations (DCT) to attain the desired rescale invariant and intensity invariant CMF detection techniques respectively. The experiments conducted quantitatively and qualitatively demonstrate the effectiveness of the algorithm.

## I. INTRODUCTION

Digital media such as images and videos serve as one of the prime mediums of information carriers. They are also well supported as evidence during law enforcement. However, in the recent year's image tampering techniques have escalated and are advanced in nature which has made us think twice about the authenticity of any image we come across. Powerful image tampering tools such as Photoshop have made image tampering easier and easier. The duplication in these images can be detected by various forgery detection mechanisms. These forgery detection mechanisms can be broadly classified into two categories namely Active methods and Passive methods. Active methods refer to the detection of hidden information inside a digital image in the form of watermarking or signatures where the embedded information can be used to identify the source of such an image or detect potential forgery in that image. Passive forgery detection methods use traces left by the post-processing steps in different phases of digital image acquisition. Passive methods mostly use the binary information in the image to find any possible traces of tampering in it and hence they do not need prior channel distribution information or the source of the digital image.

One of the most common form of passive image tampering techniques is copy move forgery where a cluster of pixels are copied from a region and are pasted in the same image in another region. An example of CMF is shown in Fig.1 (a). This kind of forgery can be detected by implementing the algorithm delineated in Fig.2 where an image of size MxN is divided into overlapping blocks of size BxB and each block is further subjected to feature extraction. Feature extraction is done with anyone of the various parameters such as DCT, DWT, DyWT, SVD-SWT, PCA, SIFT, BRISK, FREAK and many more feature extraction algorithms based on the need.

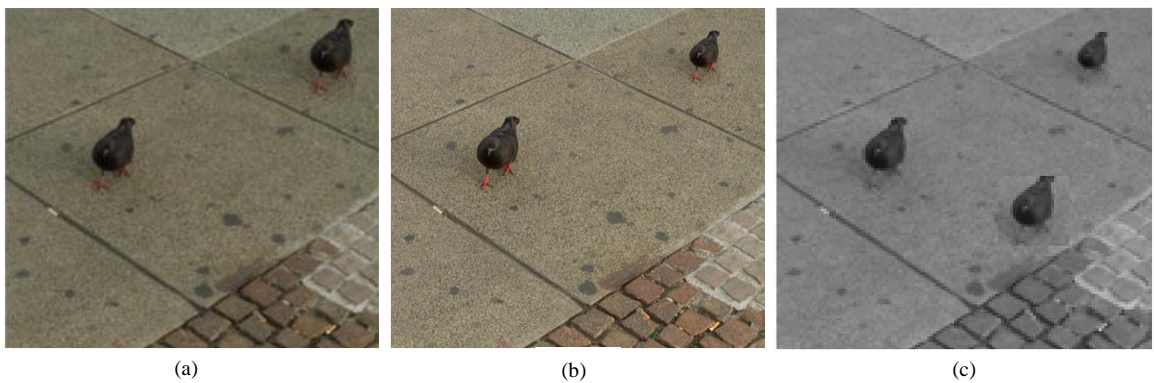

(a)                    (b)                    (c)

Fig.1. Example of (a) Copy Move Forged Image (b) Copy Rescale Moved Image (c) Intensity Varied Copy Moved Image.

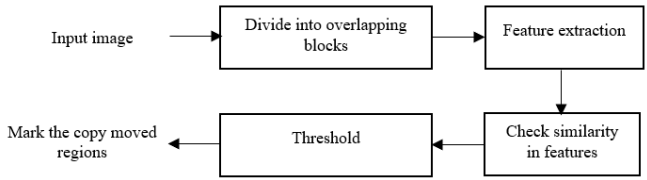

Fig2. Generalized Block Diagram of CMF detection

These extracted features are stored as linear rows of another matrix whose size equates to [(M-B)*(N-B)] x N. An additional two columns are added to this matrix which stores the location of the first pixel of the corresponding block. This feature matrix is lexicographically sorted and further features in adjacent rows are compared to find whether duplication exists. Measures are taken to further reduce the false positives by applying threshold or by calculating Euclidean distance between each block location in the copied region to the duplicated region. During this process the assumption made is that the duplication region is bigger than the block size B x B.

However, this algorithm does not hold good to identify duplication in copy – tamper –move forgeries. Here, a cluster of pixels are copied from a region and are tampered before pasting them in the same image which changes the pixel values in the duplicated region which in most cases fail to be detected by the algorithm presented in Fig.2. Some of the most common tampering techniques before placing the forged part in the source image are blurring, rotating, rescaling and varying intensity or brightness of the duplicated region. These techniques change the value of the pixels and dimensions of the matrix from the duplicated region as compared to the original region. This produces de-synchronization in the pixels which hinders the task of detecting the forgery. Among these, Blur-invariant CMF detection [19] and Rotation-invariant CMF detection [14] were made possible using SVD-SWT algorithms and Hu's invariant moments respectively. However, majority of the algorithms failed to identify rescale invariant and intensity invariant CMF detection. Therefore, to address the above mentioned limitations we propose disparate novel algorithms to detect Rescale-invariant and Intensity-invariant CMF using Hu's Invariant Moments and Discreet Cosine Transformation coefficients respectively.

## II. LITERATURE REVIEW

The passive image forgery detection technique was first introduced by Fridrich et al [1] which proposed an algorithm to detect matching parts of an image through DCT coefficients. Wang et al [2] proposed an algorithm in which a combined effect of both DCT and DWT were used. Here, mean and variance of multiplied coefficients were used which produced results robust to JPEG compression. The computational complexity was reduced by using grouped DCT coefficients was proposed by Hu et al [3]. Cao et al [4] proposed an algorithm to use Euclidian distance between blocks with matching DCT coefficients, through which it was made possible to reduce false positives drastically. Huang et al [5] worked on increasing the computational speed of DCT based CMF detection technique. Later, Myna et al [6] proposed a method which uses DWT and log polar coordinates to match the block features. Bravo and Nandi [7] adapted log polar map to propose their CMF detection of 1-D reflection and rotation invariant descriptors. Bayram et al [8] reduced computational complexity in CMF detection using Fourier Mellin Transform (FMT) and log-polar transformations. Additionally, Li and Yu [9] made the algorithm robust towards post processing operations. The algorithm was robust to slight scaling (10%) and rotation (up to 90 degrees). The use of log polar transformations was replaced with texture and intensity based algorithms. The early work in this category was done was performed by Langille et al [10] who proposed an algorithm to search for blocks with similar intensity patterns using kd-trees to address computational complexity. Xiong et al[11] proposed an algorithm which that used radix sort to reduce complexity. In order to increase robustness against post-processing such as local geometric transformations SIFT features were used by Huang et al [12]. Jin et al[13] improved it to detect multiple forgeries and also forgeries involving slight distortion. This method was robust against JPEG compression and Gaussian noise. Further, invariant key point CMF detection technique was developed, Guangjie Liu et al[14] proposed an algorithm which is robust against rotation at any degree. He used circular block division and Hu's invariant moments to achieve the robustness. The algorithm also proves efficient robustness against flipping, reflection, JPEG compression, blurring and rotation. Ryu et al [15] proposed a novel algorithm which uses Zernike moments to achieve robustness against combined post processing operations. Principal component analysis (PCA) was used to reduce feature vector size which further decreases complexity. Popescu and Farid [16] proposed the PCA based CMFD algorithm. This was further developed by K Sunil et al [17] to increase its robustness to JPEG compression and noise. Singular value decomposition (SVD) is another method to extract algebraic and geometric features from an image. Huang et al [18] proposed an algorithm using DWT and SVD for robust feature extraction. Further, Rahul et al [19] has improved the algorithm to achieve blur invariant CMFD using SWT-SVD. Kumar et al [20] have proposed an algorithm which has achieved robustness against contrast using binary DCT vectors. While the above discussed algorithms are robust against many post and intermediate processing techniques such as blurring, rotation, contrast, reflection, JPEG compression and Gaussian noise, we propose novel algorithms which could effectively achieve the desired robustness against scaling and intensity variation. Rescaled CMF is detected using rescale invariant features developed in the proposed algorithm and verified using Hu's Invariant moments. Further, we developed an exact match technique to detect intensity varied CMF using pixel block representation. Later, these blocks were replaced with DCT coefficients (only DC coefficients) to attain robustness against post-processing techniques inclusive of intensity varied CMF.

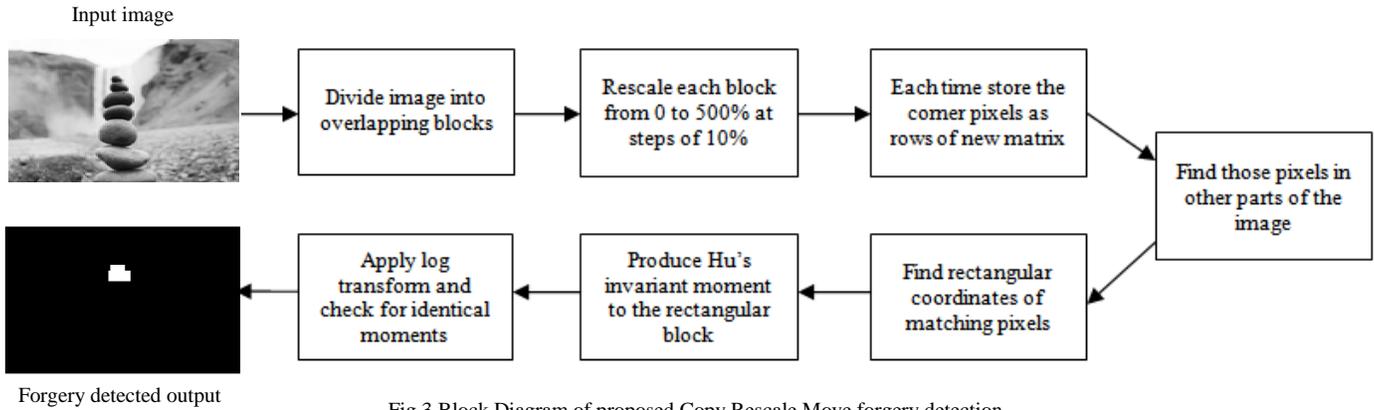

Input image

Forgery detected output

Fig.3 Block Diagram of proposed Copy Rescale Move forgery detection.

## III. PROPOSED METHOD

Passive image authentication schemes to detect copy rescale move forgery and intensity varied CMF are lucidly discussed below.

### A. Rescale Invariant Copy Move Forgery Detection.

In Rescale Invariant CMF, a part of the image is copied and is further subjected to scaling on a particular rescale factor before pasting it in another region of the source image. In majority of the cases all the pixel values in the copied region are altered in an irregular fashion and the matrix dimension of the copied and pasted regions differ depending upon the degree of scaling factor. To tackle this problem, we propose Algorithm 1, illustrated in Fig3.

---

**Algorithm 1:**

---

Input: Copy rescale move forged image.
Output: Binary image with indication of the copied part and forged part.

1. Input the forged image of size M x N.
2. Divide the image into overlapping blocks of size B x B (preferably 4x4).
3. Rescale each block from a range of 0 to 500% at steps of 10%. At each stage note the corner pixels of the resized block.
4. The corner pixels at each stage of rescaling in step 3 are made as rows of a new matrix. Also, record the rescale factor responsible for the corresponding corner pixels and add them as a new column in the vector.
5. Repeat steps 3 and 4 to each block formed in step 2 and store the matrix obtained in step 4 in unique cells of another vector.
6. Round off the all the pixel values recorded in step 5 to 15th decimal.
7. Compare the pixel values of the vector formed in Step 6 with the original pixel values of the input

image up to 15th decimal place and check for similar pixels.

8. If no similarity found in step 7, declare that image is not subjected to copy rescale move forgery.
9. If similarity found, check for the values with corner pixels in corresponding locations with similar rescale factor for all four corners in order to limit the number of false positives.
10. If four pixels of the same rescale factor are found in the image, extract the rectangular region with formed with the four pixels in respective corners.
11. Extract the corresponding copied region based on the block responsible to the values detected in step 10.
12. Produce Hu's invariant moments to the matrices extracted in step 10 and step 11 individually. Apply log polar transform for the Hu's invariant moments obtained.
13. Compare the results obtained in Step12. If results are similar mark the copy-rescale moved region's in the image.
14. If results are dissimilar in step 12, check for another set in step 9 and repeat steps 10-13. If there are no matches found, declare that input image is not subjected to copy-rescale-move forgery.

---

The copied and the duplicated portion of the image were extracted and various post processing techniques were performed to detect the source region and rescaled region in the forged image. An interesting observation was made to match the pixels between copied and the moved regions. The results obtained after rescaling each of the 4 x 4 divided overlapping blocks in the copied region, on subjecting to a particular rescale factor showed similarity in their corner pixels as compared to the result obtained when the entire copied region was rescaled with the same factor. The values of these pixels were similar to each other up to their respective 15th decimal which could reduce the number of false positives in searching the duplicated region. This property was used to generate the corner pixels and use them as features to match the forged region. These matched pixels might not be true in all cases. Hence, in order to reduce the number of false positives and to accurately estimate the forged region we apply Hu's moments which are invariant towards rescale. Matching Hu's moments would prove the authenticity in detecting the forged region.

### 1) Hu's invariant moments:

Hu's invariant moments are used to eliminate all the false positives and thereby produce results precisely as explained in Algorithm 1. Moment invariants have been extensively used to trace image patterns in tampering techniques such as image translation, JPEG compression, scaling and rotation.

### 2) Moment Invariants

Two dimensional $(p + q)^{th}$ order moment are defined as follows:

$$M_{pq} = \iint_{-\infty}^{\infty} x^p y^q f(x, y) dx dy \qquad (1)$$

Here, the function of the image f(x, y) is a continuous bounded function where the moments of all orders exist. The moment sequence $\{M_{pq}\}$ can be uniquely determined by f(x, y) and vice versa. These moments illustrated in (1) may not be invariant towards post-processing techniques. Invariant moments can be achieved through central moments, which are defined as follows:

$$\mu_{pq} = \iint_{-\infty}^{\infty} (x - \bar{x})^p (y - \bar{y})^q f(x, y) dx dy \qquad (2)$$

$(\bar{x}, \bar{y})$ represents the centroid of the image f(x, y). The centroid moments $\mu_{pq}$ are similar to $M_{pq}$ whose center is shifted to centroid of the image. This feature makes centroid moments invariant towards post processing techniques such as rotation and translation. Scale invariance can be obtained through normalization of central moments which is defined as follows:

$$\eta_{pq} = \frac{\mu_{pq}}{\mu_{00}^{\gamma}}, \gamma = \frac{p + q + 2}{2} \qquad (3)$$

Based on normalized central moments Hu introduced seven moment invariants out of which we use four distinguished moments in order to reduce the dimension of feature vector:

$$\emptyset_1 = \eta_{20} + \eta_{02} \qquad (4)$$
$$\emptyset_2 = (\eta_{20} - \eta_{02})^2 + 4\eta_{11}^2$$
$$\emptyset_3 = (\eta_{30} - 3\eta_{12})^2 + (3\eta_{21} - \mu_{03})^2$$
$$\emptyset_4 = (\eta_{30} + \eta_{12})^2 + (\eta_{21} + \mu_{03})^2$$

These moment invariants introduced by Hu are invariant towards translation, JPEG compression, rotation and rescale.

### B. Intensity Invariant copy move forgery detection.

Intensity varied copy move forgery refers to the process of adding a certain constant to the copied region before forging it into the image. This constant added might be different to different blocks of copied region which creates a de-synchronization in the searching process failing to detect forgery. To solve this problem, two dissimilar solutions are proposed in the paper, the first aiming to secure the best accuracy and another aiming to execute faster. Algorithm 2 is well explained by the flowchart in Fig.4.

---

**Algorithm 2:**

---

Input:   Intensity varied CMF image.
Output:  Forgery detected and marked in the input image.

1. Input the forged image of size M x N, convert it to grey scale.
2. Divide the image into overlapping blocks of size B x B.
3. Subtract each Block with every other block.
4. If all the elements in the result of step 3 consists of same value greater than zero, mark the regions of both the matrices as copied and moved respectively.
5. Check for pixel value = 255 in the block subjected in step 3.
6. If a pixel=255, the corresponding location of the matrix obtained in step 3 is swapped with the maximum value of pixel within the matrix.
7. Now recheck if all elements in the matrix are equal. If equal, mark the copy moved regions respectively.
8. If no elements are equal, output "Forgery not found".

The above algorithm proves maximum efficiency in detecting the forged part accurately, assuming that the forged region is of size greater than that of block size B x B. For better results we choose block size to be 4 x 4. The maximum value achieved by a pixel in an image is 255, which restricts the effect of adding constant to forge a region. Similarly, a value less than zero cannot be achieved by pixels in an image. These test cases were precisely considered and solved in Algorithm 2. Alternatively, Algorithm 2 can be replaced by a robust match where robust representation of blocks are used instead of pixel representation. Robust representation can be formed using DCT coefficients. The advantage of DCT is that the energy is localized to the first few coefficients. Therefore, changes in higher frequencies occurring due to intermediate and post-processing techniques such as noise addition, JPEG compression and retouching where the first few coefficients are not affected drastically.

Input image

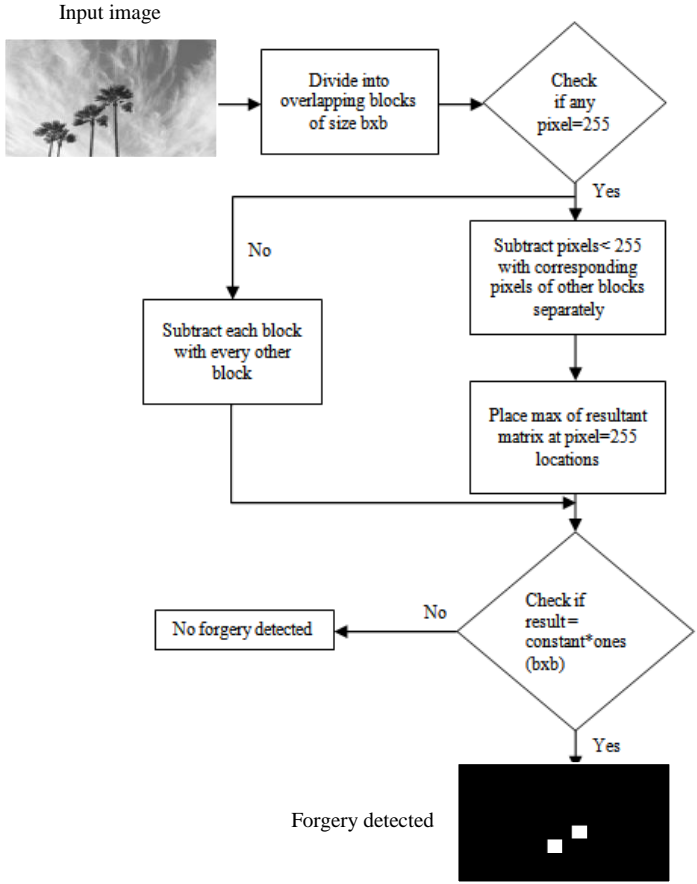

Fig.4. Flowchart describing Intensity Varied CMF

---

**Algorithm 3:**

---

Input: Intensity varied CMF image.
Output: Forgery detected and marked in the input image.

1. Input the forged image, convert it to grey scale.
2. Divide the image into overlapping B1=blocks of size B x B.
3. Produce dct1= DCT coefficients of size B x B. Compute dct1*B1*dct1$^T$ for each block.
4. Round off the values to 13$^{th}$ decimal.
5. The results from step 3 are converted into linear rows of a new matrix.
6. An additional two columns are added to the matrix formed in step 4 representing the location of their corresponding blocks.
7. The first column, representing the DC coefficients of the blocks is deleted.
8. Lexicographically sort the matrix.
9. Check for adjacent equal rows in the matrix.
10. If found, find the Euclidean distance between the locations of the blocks which were found similar in Step 9.

11. Record all the Euclidean distances in an array and apply user specified threshold to eliminate false positives.
12. Blocks responsible for values above specified threshold are marked to be CMF.

---

DCT works on real numbers to express a finite sequence of data points in terms of sum of cosine functions oscillating at different frequencies. DCT has been widely used to represent the image in frequency domain where it discards both the high and low frequencies in the image and represents most of its intensity distribution details concentrated at fewer coefficients. DCT is often used due to its strong energy compaction property and de-correlation property. Two dimensional DCT of an M x N image is given by the equation:

$$c(u,v) = \alpha(u)\alpha(v) \sum_{X=0}^{N-1} \sum_{U=0}^{M-1} f(x,y) \cos\frac{\pi(2x+1)u}{2N} \cos\frac{\pi(2y+1)v}{2M}$$

(5)

where $\alpha(i) = \begin{cases} \sqrt{1/N} \ for \ i=0 \\ \sqrt{2/N} \ for \ i \neq 0 \end{cases}$

u=0, 1, 2……N-1 and v=0, 1, 2….M-1

The below function is called basis function of DCT.

$$\alpha(u)\alpha(v) \cos\left(\frac{\pi(2x+1)u}{2N}\right)\cos\left(\frac{\pi(2y+1)v}{2M}\right)$$

(6)

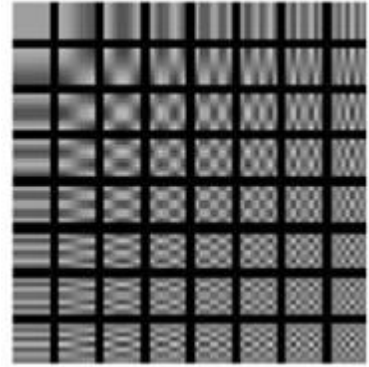

Fig.5. DCT Basis Function of an 8x8 Block

Considering the block size to be 8 x 8, 64 basis functions can be formed as shown in the Fig. 5. Frequencies are distributed in ascending order from left to right and vertically from top to bottom. This proves the property of energy localization among DCT coefficients. The first DCT co efficient represents the average intensity over the particular block chosen in the image and is called as DC coefficient. The other coefficients represent variations in the intensity and are termed as AC coefficients. Therefore, when a constant is added to the block, the DCT coefficients of the resultant are found to be the same up to 13$^{th}$ decimal except the first element i.e. AC coefficient values did not differ whereas DC coefficient value changed depending on the constant added. These AC coefficients are brightness or intensity invariant. However, they are not contrast invariant. This property was observed and further used to predict intensity varied CMF as illustrated in Algorithm 3. Block diagram in Fig 6 illustrates the flow in Algorithm 3.

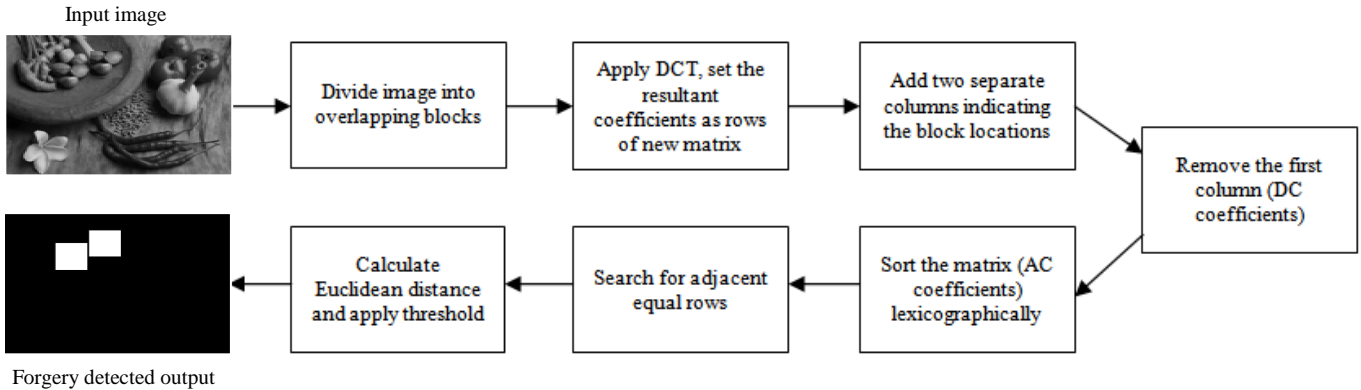

Fig.6. Block Diagram of Intensity Varied CMF Detection using DCT coefficients.

## IV. EXPERIMENTAL RESULTS

A set of 315 images from the official database MICC – F220 and other grey scale images which were manually forged were chosen for the experimental analysis. 105 images were tested on each algorithm and corresponding observations were recorded. For color images, the detection can be performed on intension components.

We chose standard size of the image to be 256 x 256. However, overlapping square window size varied from 8 x 8 to 4 x 4, in order to attain the best results possible.

### A. Robustness against rescale

We manually forged a set of 105 images where the copied region was subjected to dissimilar scaling factors before moving to another region. When these results were passed through Algorithm 1, implemented in MATLAB 2016a software, the results illustrated in Fig 7 were achieved. Here, Fig 7(a) represents the original image, Fig 7(b) represents the copy rescale move forged image and Fig 7(c) delineates the detection results obtained of the copied and detected region.

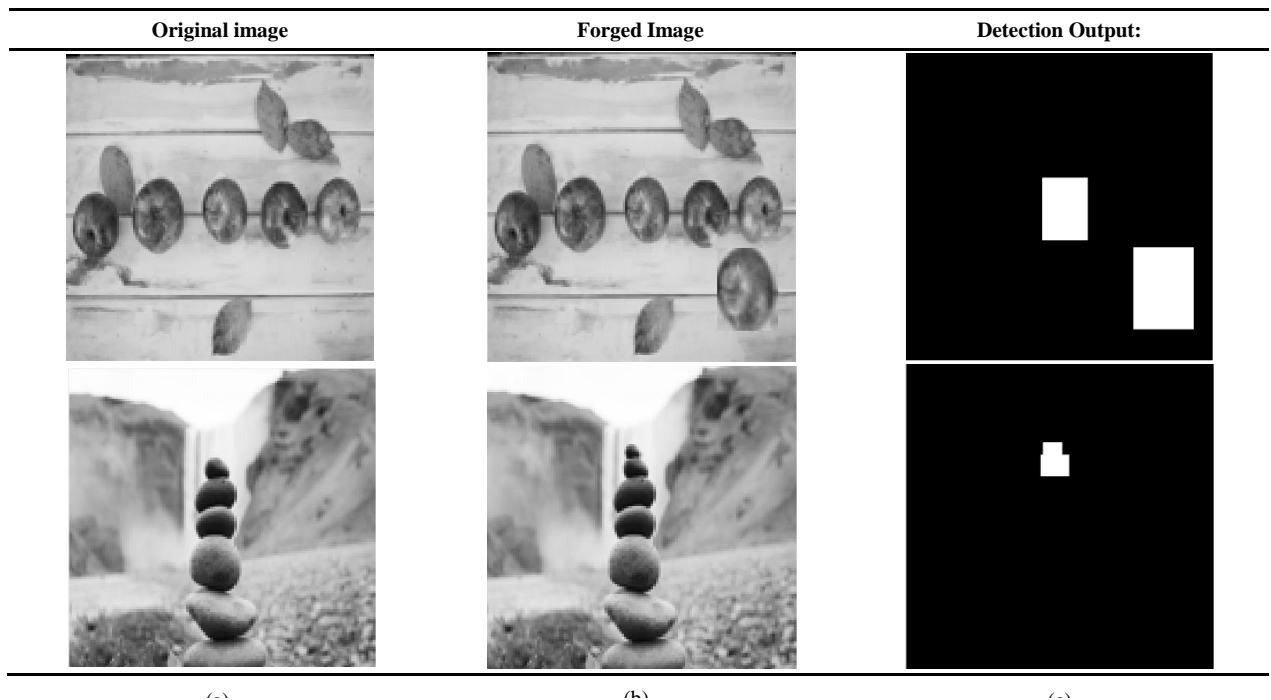

Fig.7. Results obtained on testing Algorithm 1. (a) Original image. (b) Copy Rescale Moved Image. (c) Forgery Detection Results

Table 1: Invariant moments after log transform at different scaling factors.

| Hu's invariant Moment absolute log value/ scaling factor | Actual image 100 % | 50% | 170% | 250% | 300% | 500% |
|---|---|---|---|---|---|---|
| N1 | 0.5334 | 0.5504 | 0.5120 | 0.5237 | 0.5231 | 0.5223 |
| N2 | 2.8864 | 2.8863 | 2.8589 | 2.8574 | 2.8589 | 2.8559 |
| N3 | 4.6126 | 4.5402 | 4.5636 | 4.5731 | 4.5636 | 4.5584 |
| N4 | 4.1340 | 4.1208 | 4.1322 | 4.1326 | 4.1322 | 4.1319 |

In order to evaluate its performance, the copied region was subjected to a scaling factor of 0 to 500% in multiples of 10. To check on the robustness of Hu's invariant moments in detecting the forgery at various scaling factors a test was performed. A random block from the image in Fig 7(a), pixels (15:22, 15:22) was extracted whose size was 8 x 8 and was subjected to different scaling factors. Hu's invariant moments up to fourth order were obtained for each of the scaled image. As the order of the invariant moments increase, the value of the moments generally tends to go beyond $10^{-6}$. Therefore, in order to visualize the results better we apply log polar transform to the invariant moments obtained at each order. The results obtained are illustrated in Table-1.

In Table-1 we can observe that the invariant moments do not differ in a significant manner. This property can be used to verify the copy rescale move forgery detection. Another advantage of using Hu's invariant moments as compared to other alternative feature extraction methods [3- 20] is that the feature dimension size is comparatively lesser. Hu's moments tend to reach beyond $10^{-6}$ after fourth moment and hence we could consider moments up to fourth order for valuable reasoning.

Table 2: Feature Vector Dimension

| Methods | Extraction domain | Block amount | Feature Dimension. |
|---|---|---|---|
| Farid et al [16] | PCA | 14,641 | 32 |
| Myna et al [6] | DWT/DyWT | 14,641 | 64 |
| Fridrich et al [3] | DCT | 14,641 | 64 |
| Guangjie Liu [14]/ Proposed Algorithm | Hu | 14,641 | 4 |

### B. Robustness against intensity variation:

The images were manually forged in a way the blocks moved had a constant difference in them. This can be done either by creating multiple intensity differences between each other or the same intensity for all duplicated regions. The assumption in this detection technique is that the forgery is not smaller than the block size B x B. Algorithm 2 provides the best accuracy whereas Algorithm 3 is more robust towards post-processing techniques. The results obtained from both algorithms are illustrated in Fig. 9 and Fig. 10 respectively. These algorithms also prove effective to detect multiple forgeries formed from cloning. The inter-performance variation can be observed from Table-3.

### C. Performance evaluation [19]

The quantitative measurements of the performance (P) of proposed algorithms were evaluated by the formula:

$$P = \frac{\text{Number of correctly detected copy} - \text{moved pixels}}{\text{Number of pixels actually copy- moved}} \times 100 \%$$

Table 3: Performance Evaluation

| Method | Forgery size 10% (P in %) | Forgery size 20% (P in %) | Forgery size 30% (P in %) | Forgery size 40% (P in %) | Average (P in %) |
|---|---|---|---|---|---|
| Proposed Algorithm1 | 98.8863 | 98.6541 | 98.4545 | 97.9161 | 98.4777 |
| Proposed Algorithm 2 | 100 | 100 | 100 | 100 | 100 |
| Proposed Algorithm 3 | 97.8672 | 97.4396 | 97.6434 | 97.0624 | 97.7531 |

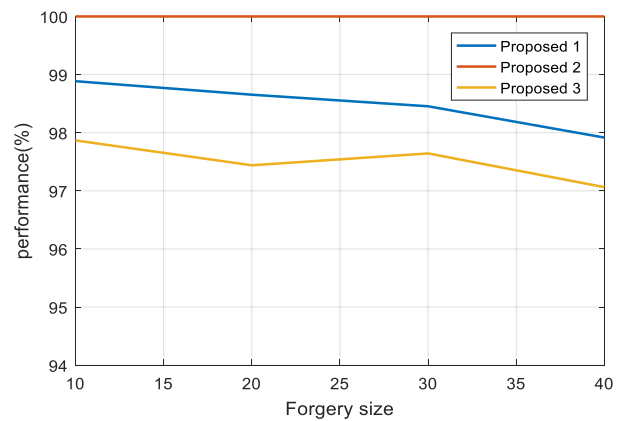

Fig.8. Variation of Performance with forgery size for different proposed algorithms

A unique set of 105 images were tested on each of the proposed Algorithms. These images included forgeries with a broad range of duplication size. The average performance (P) results of each case have been demonstrated in Table 3 and Fig.8. Table-4 illustrates various state-of-art techniques and their respective capabilities. It provides a comprehendible comparison of the capabilities of other algorithms in detecting various post processing techniques with the proposed algorithms.

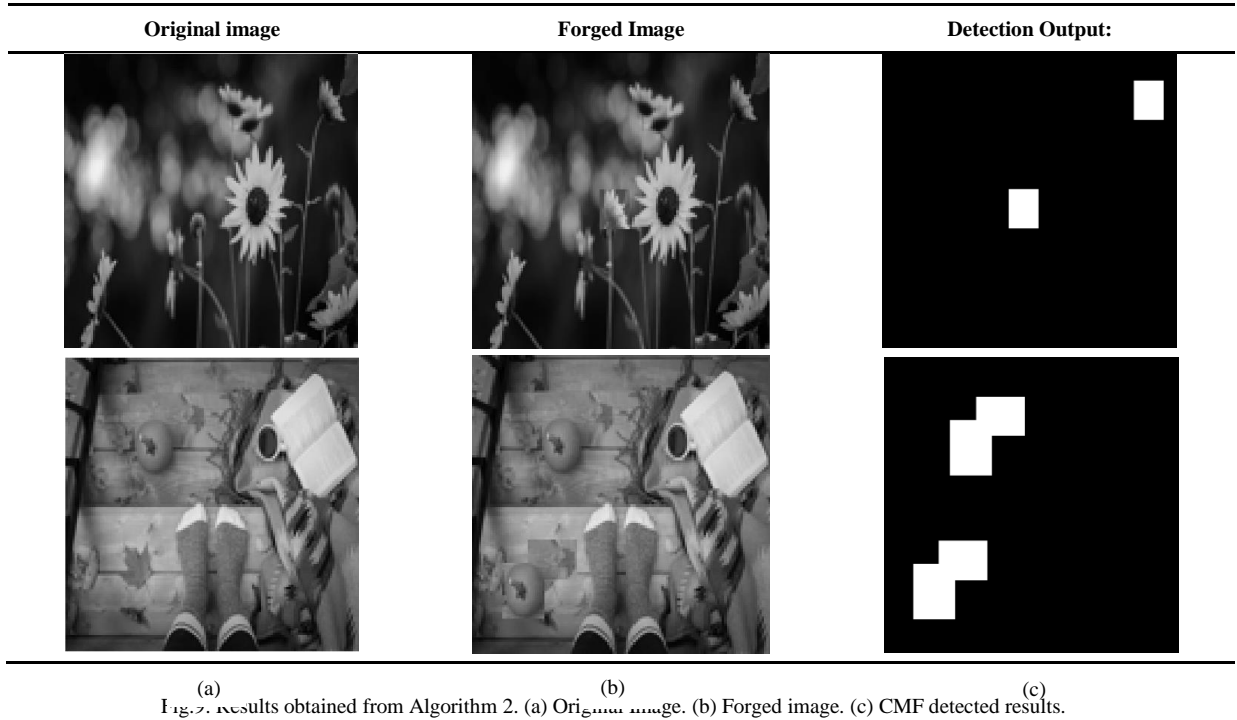

| Original image | Forged Image | Detection Output: |
|:---:|:---:|:---:|
| (a) | (b) | (c) |

Fig.9. Results obtained from Algorithm 2. (a) Original image. (b) Forged image. (c) CMF detected results.

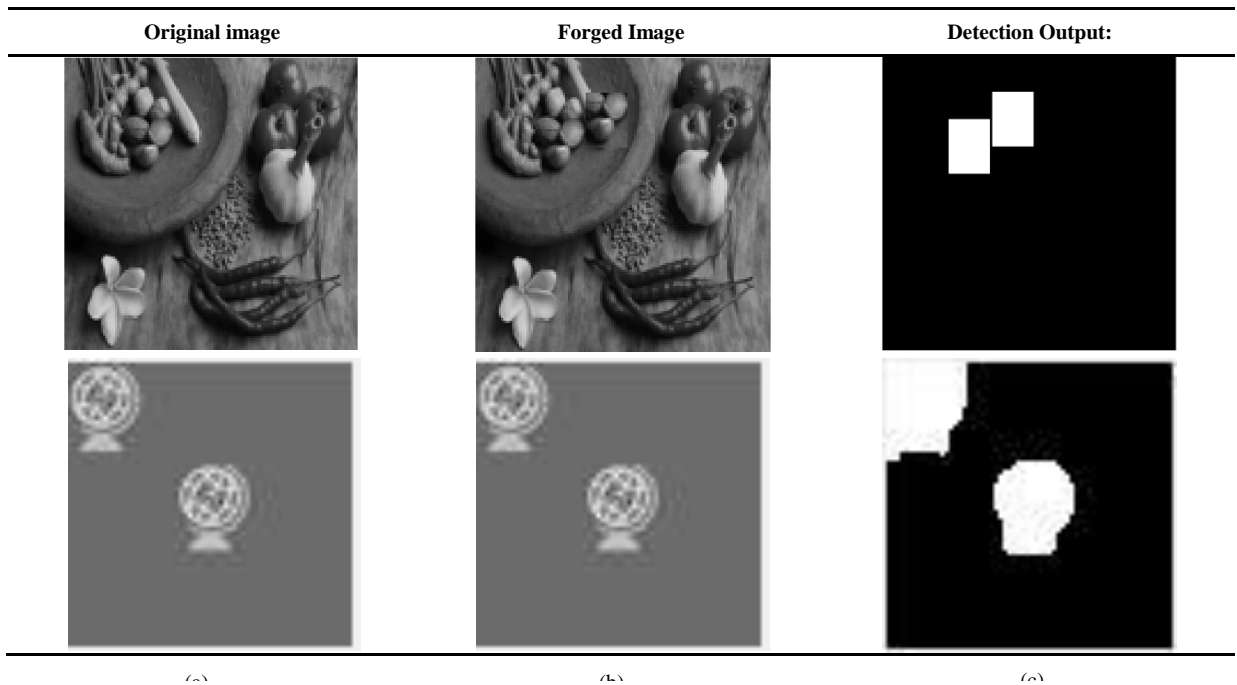

| Original image | Forged Image | Detection Output: |
|:---:|:---:|:---:|
| (a) | (b) | (c) |

Fig.10. Results obtained from Algorithm 3. (a) Original Image. (b) Forged image. (c) CMF detected results.

Table 4: Comparison of capabilities of the proposed algorithms with the existing state of art techniques.

| Feature Extraction Method. | Authors | Reflection | Scaling | JPEG compression | Additive Gaussian Noise | Intensity |
|---|---|---|---|---|---|---|
| DCT | [1-5] | ✘ | ✘ | ✔ | ✔ | ✘ |
| Invariant points | [12-13] | ✔ | ±10% | ✔ | ✔ | ✘ |
| Invariant moments | [14-15] | ✔ | ✘ | ✔ | ✔ | ✘ |
| PCA | [16-17] | ✘ | ✘ | ✔ | ✔ | ✘ |
| SWT-SVD | [19] | ✘ | ✘ | ✔ | ✔ | ✘ |
| Hu's invariant moment | Proposed Algorithm 1 | ✔ | ✔ | ✔ | ✔ | ✘ |
| Exhaustive search | Proposed Algorithm 2 | ✔ | ✘ | ✘ | ✘ | ✔ |
| DCT | Proposed Algorithm 3 | ✘ | ✘ | ✔ | ✔ | ✔ |

## V. CONCLUSION

Applying intermediate and post processing techniques to the duplicated region while creating a CMF is a common strategy to avoid detection. Adding intensity or scaling the duplicated region were some of the common forms of tampering the duplicated region to create de-synchronization in the searching process. To tackle this, novel solutions using Discreet Cosine Transformations and Hu's invariant moments were provided, which proved to be robust against adding intensity and scaling respectively. In the future, work can be focused on developing a single algorithm that could be robust towards multiple intermediate and post processing techniques.